\pdfoutput=1

\documentclass[11pt]{article}
\usepackage{pgfplots}
\usepackage[final]{coling}

\usepackage{times}
\usepackage{latexsym}

\usepackage[T1]{fontenc}
\usepackage{tcolorbox}
\usepackage[utf8]{inputenc}
\usepackage{tikz}
\usetikzlibrary{patterns}
\usepackage{microtype}

\usepackage{inconsolata}

\usepackage{graphicx}

\usepackage{microtype}
\usepackage{booktabs}
\usepackage{amssymb}
\usepackage{hyperref}
\usepackage{amsmath}
\usepackage{subfig}
\usepackage{graphicx}
\usepackage{multirow}

\title{Enabling Real-Time Conversations with Minimal Training Costs}

\author{Wang Xu$^{1}$, Shuo Wang$^{1}$, Weilin Zhao$^{1}$, Xu Han$^{1}$, Yukun Yan$^{1}$, Yudi Zhang$^{2}$, Zhe Tao$^{2}$ \\
\textbf{Zhiyuan Liu$^{1}$, Wanxiang Che$^{2}$\thanks{Corresponding author: Wanxiang Che}}\\
{$^{1}$Department of Computer Science \& Technology, Tsinghua University, Beijing, China} \\
{$^{2}$Harbin Institute of Technology, Harbin, China}\\
\texttt{xwjim812@gmail.com, car@ir.hit.edu.cn},
}

\begin{document}
\maketitle
\begin{abstract}
Large language models (LLMs) have demonstrated the ability to improve human efficiency through conversational interactions.
Conventional LLM-powered dialogue systems, operating on a turn-based paradigm, preclude real-time interaction during response generation. 
To address this limitation, researchers have proposed duplex models. 
These models can dynamically adapt to user input, facilitating real-time interactive feedback.
However, these methods typically require substantial computational resources to acquire the ability.
To reduce overhead, this paper presents a new duplex decoding approach that enhances LLMs with duplex ability, requiring minimal additional training.
Specifically, our method employs parallel decoding of queries and responses in conversations, effectively implementing a channel-division-multiplexing decoding strategy.
Experimental results indicate that our proposed method significantly enhances the naturalness and human-likeness of user-AI interactions with minimal training costs.
\end{abstract}

\section{Introduction}
\label{sec1}
LLMs have revolutionized human-computer interaction, enabling applications like question answering~\cite{zhu-etal-2024-fanoutqa,chatgpt,zhao2023surveylargelanguagemodels} and coding assistance~\cite{Rozire2023CodeLO,han-etal-2024-archcode} that augment human capabilities through conversation. 
Consequently, the quality of these interaction experience is important.

Traditional turn-based chat systems inherently limit the interactive experience~\cite{hill2015real,zhou2023talking,zhang2024turnbasedgameenablingrealtime}.
Current human-AI interactions follow a turn-based pattern, with one party passively waiting while the other responds~\cite{skantze2021turn}.
Interruptions are facilitated through manual interventions, such as a "stop" function, resulting in communication that lacks fluidity.
However, human conversations involve simultaneous listening and thinking. 

Recent studies have proposed duplex models to address this challenge~\cite{zhang2024turnbasedgameenablingrealtime,fang2024llamaomniseamlessspeechinteraction,fu2024vitaopensourceinteractiveomni}. 
\citet{zhang2024turnbasedgameenablingrealtime} introduced MiniCPM-Duplex, a novel approach that addresses the duplex modeling challenge. 
This method employs a time-division-multiplexing strategy for encoding and decoding.
The input and output are split and mixed in a time slice format, enabling pseudo-simultaneous processing of text segments.
Additionally, \citet{ma2024languagemodellistenspeaking} introduced LSLM which combines channels for autoregressive generation and real-time turn-taking detection. 
Three strategies are explored to fuse the listening and speaking channels.

\begin{figure*}[t!]
  \includegraphics{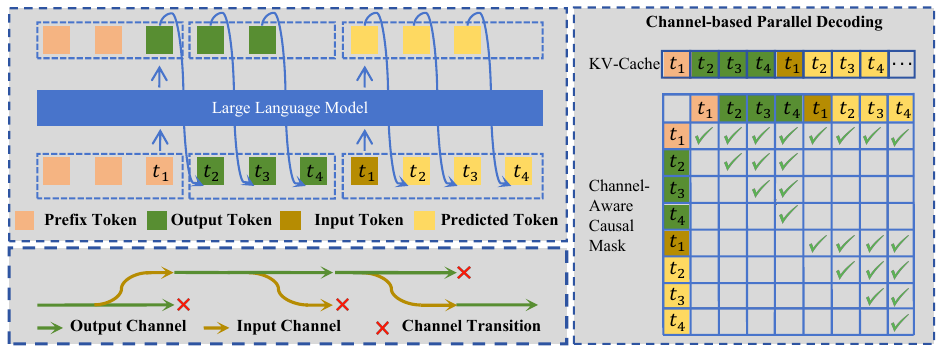}
  \caption {\label{fig.duplex}\textbf{Top Left}: A new decoding branch is established when a user interprets the model's generation. DUO doesn't increase the forward number compared to the standard decoding. \textbf{Right}: The tokens generated by the input and output channels after time step $t_1$ do not attend to each other, despite sharing the same prefix tokens. \textbf{Left Bottom}: Channel transition is activated when the state tokens are predicted. }\vspace{-2mm}
\end{figure*}
However, these approaches often demand significant computational resources for model retraining, as the designed procedure diverges substantially from the backbone model.
To address this issue, we present a \textbf{DU}plex dec\textbf{O}ding~(\textbf{DUO}) approach, a novel channel-based parallel decoding mechanism that equips LLMs with duplex capabilities.
Our approach enables simultaneous autoregressive output generation and input preprocessing.
Furthermore, two state tokens are designed to signal whether a query should be addressed or ignored.
The proposed method requires only minimal additional training to learn the state of the query.

To demonstrate the effectiveness of the proposed method, we test DUO based on human evaluation and standard benchmarks.
The results show that our method significantly enhances the naturalness and human-likeness of user-AI interactions with minimal training costs.

\section{Methodology}
\label{sec2}
We consider two scenarios: \textbf{non-awakening interaction} and \textbf{interruption interaction}~\cite{fu2024vitaopensourceinteractiveomni}. 
Specifically, the model should filter out background dialogues or non-query input.
Furthermore, if the user interrupts with another question during an ongoing generation, the model should pause its current output to immediately address the latest query.

In this section, we introduce DUO in detail.
DUO processes inputs and outputs cyclically.
In each cycle, the output channel generates new tokens autoregressively, while the input channel prefills the key-value cache and predicts the next tokens. 
According to the predicted tokens, the model decides whether a query is addressed or ignored.

\subsection{Parallel Decoding}
\label{sec2-1}
As illustrated in Figure~\ref{fig.duplex}, a new input token is received at time step $t_1$ while the model generates output autoregressively, thus establishing a new input channel.
Subsequently, the model generates tokens autoregressively while simultaneously processing the input tokens.

Note that the tokens generated by the input and output channels after $t_1$ do not attend to each other, despite sharing the same prefix tokens. This preserves language modeling in each channel. 
To accommodate the token dependency, the attention mask is modified as shown on the right of Figure~\ref{fig.duplex}.

Upon processing the input tokens, the output channel persists in generating tokens rather than terminating immediately. 
This output channel is maintained until either the model responds to the new query or the output sequence is completed. 
This capability is crucial in scenarios where the assistant continues to generate responses even when non-query input is received.

\subsection{Channel Transition}
\label{sec2-2}

It is crucial for the model to determine both whether and when to respond to new input. 
To address this issue, we use state tokens to indicate the status of the user's query~\cite{fu2024vitaopensourceinteractiveomni}. 
Specifically, the model predicts $\alpha$ tokens to complete the query in each processing cycle. 
Upon encountering a state token, the model would handle different interactive behaviors as follows:
State token \textbf{\textit{<1>}} denotes that the input is a user query. The original output channel is suspended, and the input channel seamlessly transitions to become the new output channel.
State token \textbf{\textit{<2>}} denotes that the new input is no-query text. 
Thus the model continues to respond to the original query, and the input channel is disregarded.
Otherwise, the key-value cache of the predicted would be disregarded and the next processing cycle goes on.

Our method DUO differs from MiniCPM-Duplex~\cite{zhang2024turnbasedgameenablingrealtime}. 
Essentially, MiniCPM-Duplex processes the input and generates output sequentially, functioning as a time-division-multiplexing system.
To adapt to time-division-multiplexing, MiniCPM-Duplex splits and mixes the input and output in a time slice format.
This change means it needs a lot of computing power to retrain the model, as it's quite different from the original training data. 
In contrast, DUO processes the input and output simultaneously by optimizing the decoding strategy, functioning as a channel-division multiplexing system.
The data format remains largely unchanged, with only the addition of some state tokens, thereby preserving the ability of the original model.
The training example of MiniCPM-Duplex and our DUO is shown in Figure~\ref{fig.data_duplex} and Figure~\ref{fig.data_duo}, respectively.
Moreover, DUO is capable of handling non-awakening interactions, an aspect that MiniCPM-Duplex does not address.

DUO can be efficiently implemented using FlexAttention~\cite{flexatten}, which facilitates non-standard attention mechanisms.
The operations are all accomplished through the management of the key-value cache.
Importantly, DUO does not increase the number of forward passes compared to standard decoding. 
As illustrated in Figure~\ref{fig.duplex}, DUO processes the input and output channel in each time slice. 
The model generates output concurrently while the key-value cache of input is prefilled and the next tokens of input are predicted.

\subsection{Dataset Construction}
\label{sec2-3}
We introduce two state tokens to determine whether to respond to the query.
This section details our process for constructing the dataset from Duplex-UltraChat~\cite{zhang2024turnbasedgameenablingrealtime} that structures conversations in a time-division format.
Each individual may interrupt before the other participants' generation is completed and transition sentences are added to enhance the interactive experience.

We have adapted Duplex-UltraChat into a standard turn-based format.
As previously outlined, we append state tokens to indicate whether to respond to the query. 
Since all queries in the dataset require a response, we append the \textit{<1>} token to each user's content.
To simulate non-awakening interaction scene, we use ChatGPT to generate the non-query text.
We optimize the prompt to make the non-query text relevant to the context.
The prompt is described in Appendix~\ref{app_a}.

\section{Experiments}
\label{sec3}

\subsection{Setup}
\label{sec3-1}

Following \citet{zhang2024turnbasedgameenablingrealtime}'s work, we utilize MiniCPM-2.4B~\cite{hu2024minicpm} as our foundation model and 10K samples are constructed to train the model as described in Section~\ref{sec2-3}.
It is worth noting that our dataset is significantly smaller than the one used in MiniCPM-Duplex, which consists of approximately 5,000K samples.
We use MiniCPM-Duplex as our baseline since we are all derived from the same checkpoints.

The training of MiniCPM-Duo uses the following hyperparameter:
1e-5 learning rate, a constant learning rate scheduler, a batch size of 112, and a maximum length of 4,096. 
The loss is calculated on the state tokens and output.
MiniCPM-Duo is trained for 400 steps on 8 NVIDIA A100 GPUs for about 25 minutes.
However, MiniCPM-Duplex is trained for 10,000 steps on 40 NVIDIA A100 GPUs for 36 hours.
We implement the decoding strategy based on the repository~\footnote{https://github.com/thunlp/Ouroboros}
The input channel predicted tokens $\alpha$ are set to 4.

\subsection{Main Results}
\label{sec3-2}

\begin{figure}[t!]
  \includegraphics{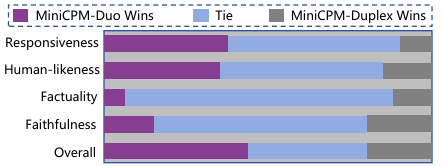}\vspace{-1mm}
  \caption {\label{fig.comparion}The comparison result between MiniCPM-Duo and MiniCPM-Duplex on responsiveness, human-likeness, factuality, faithfulness, and overall satisfaction.}
\end{figure}

\textbf{Human Evaluation.}
To evaluate the effectiveness of the proposed method, we use human evaluators. 
Following \citet{zhang2024turnbasedgameenablingrealtime}'s work, we consider four aspects: responsiveness, human-likeness, faithfulness, and factuality.
Responsiveness evaluates whether the model responds to user queries.
Human-likeness assesses how closely the model's responses resemble those of a human.
Faithfulness measures the extent to which the model adheres to user instructions~\cite{adlakha2023evaluating}.
Factuality evaluates the level of content that is not grounded in factual information~\cite{wang2023survey}.

\begin{figure*}[!t]
  \includegraphics{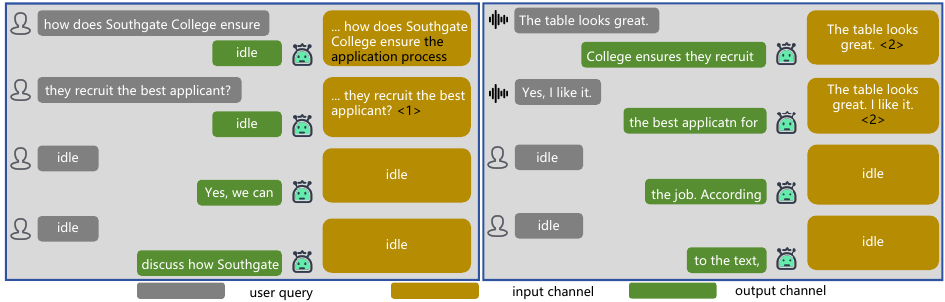}
  \caption {\label{fig.duplex_case}Case study. The black text denotes the predicted text in the input channel. }
\end{figure*}
To demonstrate our proposed model, MiniCPM-Duo, we adapted the interactive demo~\footnote{https://github.com/thunlp/duplex-model} originally developed for MiniCPM-Duplex. 
In this demo, users engage in voice-based conversations with an AI assistant. 
The speech-to-text and text-to-speech conversions are facilitated by Google's cloud ASR and TTS APIs, respectively.
We conducted 40 multi-turn dialogue sessions with each model through this demo and recorded the interactions. 
To evaluate the dialogue quality, we enlisted three participants, each holding a Bachelor's degree, to assess the dialogue histories. The evaluation criteria included: responsiveness, human-likeness, faithfulness, actuality, and overall experience.
Similarly, 
For each criterion, participants were asked to rank the dialogues generated by MiniCPM-Duo and MiniCPM-Duplex.

The comparative analysis results are presented in Figure~\ref{fig.comparion}.
MiniCPM-Duo demonstrates superior responsiveness and human-likeness compared to MiniCPM-Duplex, while maintaining comparable levels of factuality and faithfulness.
Furthermore, MiniCPM-Duo exhibits a substantial advantage in overall performance.

\begin{table}[!t]
    \centering
    \resizebox{0.95\linewidth}{!}{
    \begin{tabular}{l|cccc}
        \toprule
        \textbf{Benchmark} & \textbf{MiniCPM} & \textbf{+Duplex} & \textbf{+Duplex} & \textbf{+Duo} \\
        \midrule
        Data Used  & - & +~5,000K  & +~10K & +~10K \\
        \midrule
        CMMLU  & \textbf{51.30} & 48.53  & 50.56 & 50.66  \\
        MMLU & 53.45 & \textbf{53.76}   & 52.89 & 53.13 \\
        BBH & \textbf{37.25} & 36.35 & 35.81 & 36.51  \\
        \midrule
        HumanEval & 50.00 & 49.39 & 51.81 & \textbf{51.83}  \\
        MBPP & 38.09  &\textbf{38.30} & 34.70 & 38.19  \\
        \midrule
        GSM8K & 42.30 & \textbf{46.10}  & 40.26 & 42.15 \\
        \midrule
        ARC-e & 84.60 & \textbf{85.19} & 83.12 & 84.93  \\
        ARC-c & 69.80 & 70.05 & 69.43 & \textbf{70.48}   \\
        HellaSwag & \textbf{61.40} & 60.79 & 50.85 & 59.78  \\
        \bottomrule
    \end{tabular}
    }
    \caption{\label{tab.commonbench}Performances on standard benchmarks. Data used denotes the number of data to train MiniCPM.}
\end{table}

\textbf{Standard Benchmarks.}
We make the model capable of duplex with minimal training.
Following \citet{zhang2024turnbasedgameenablingrealtime}'s work, we evaluate our model on several standard benchmarks, including multitask (CMMLU~\citep{li2023cmmlu}, MMLU~\citep{hendrycks2020measuring}, BBH~\citep{suzgun2023challenging}), code (HumanEval~\citep{chen2021evaluating}, MBPP~\citep{austin2021program}), math (GSM8K~\citep{cobbe2021training}), and reasoning (ARC-e, ARC-c~\citep{clark2018think}, HellaSwag~\citep{zellers2019hellaswag}) with the LLM evaluation platform, UltraEval~\citep{he2024ultraeval}. 

Table~\ref{tab.commonbench} presents the primary results on general benchmarks of MiniCPM-Duo and MiniCPM-Duplex.  
MiniCPM-Duo achieves comparable results with MiniCPM-Duplex and MiniCPM.
It is noted that MiniCPM-Duplex requires substantial computational resources to learn the time slice format of input and output.
In contrast, our proposed method preserves the model's original capabilities to a greater extent.
We solely train the model to recognize the state of the user's query, a process that demands minimal computational resources.

We study how the MiniCPM-Duplex performs when the model is trained on the same scale dataset as MiniCPM-Duo.
The training hyper-parameters are the same as ours.
The results are shown in Table~\ref{tab.commonbench}, the performances drop.
Moreover, the duplex capability of inadequately trained MiniCPM-Duplex is relatively poor. 

\textbf{Case study.}
We present a case study as illustrated in Figure~\ref{fig.duplex_case}.
On the left side, a user initiates a query.
The model prefills the key-value cache and predicts the next tokens.
Upon predicting token \textit{<1>} and observing user silence, the model commences autoregressive answer generation.
On the right side, the user interrupts the generation.
The model continues generating the output while simultaneously completing the input. 
However, when the model predicts a token \textit{<2>}, the new query is interpreted as a no-query voice, resulting in the deletion of the new input.
This case study demonstrates the model's adaptive behavior in response to user interactions and silences during the query and generation processes.

\section{Conclusion}
\label{sec4}
This paper presents DUO, a novel method that equips models with duplex abilities, requiring minimal training costs.
DUO employs channel-division multiplexing, generating output autoregressively while simultaneously processing input.
The experiments show that the models achieve real-time interactive feedback, including non-awakening interaction and interrupt interaction. 

\section*{Limitations}
\label{sec5}
While DUO demonstrates promising results in equipping large language models with duplex capabilities, the study has limitations. 
A significant constraint lies in the scope of multimodal testing. The experiments primarily focused on language models, potentially limiting the generalizability of findings to other domains.
In future research, we aim to extend DUO to multimodal settings. The DUO approach is inherently designed for versatility and can be readily adapted to multimodal large models. 
In this setting, input tokens could be features extracted by encoders from various modalities, thereby further broadening the potential applications of this technology.

\bibliography{custom}

\appendix

\section{Appendix}
\label{sec:appendix}

\subsection{Data Construction}
\label{app_a}

\begin{figure}[ht!]
\begin{center}
\begin{tcolorbox}[colback=gray!5!white,colframe=gray!75!black,title=System Prompt]

The following is a multi-turn conversation.
Please review the conversation and generate coherent statements from the user’s perspective.
These sentences are created by the user and do not require a response from the assistant.
Please remember to use declarative statements, not questions.\\

\{multi-truns\} \\

Just give me your answer, no other explanation.
\end{tcolorbox}
\end{center}
\caption {\label{fig.prompt}The prompt used for data construction.}
\end{figure}

\subsection{Traing Data Example}
\label{app_b}
\begin{figure*}[t!]
  \includegraphics{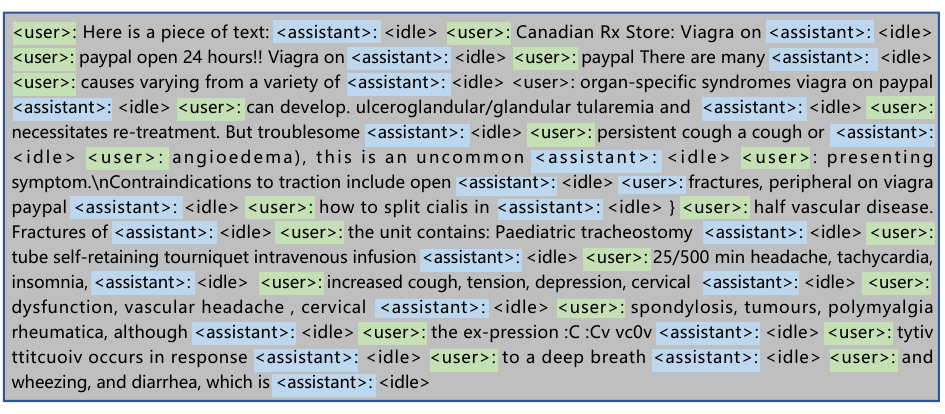}
  \caption {\label{fig.data_duplex}The training data example of MiniCPM-Duplex.}
\end{figure*}

\begin{figure*}[t!]
  \includegraphics{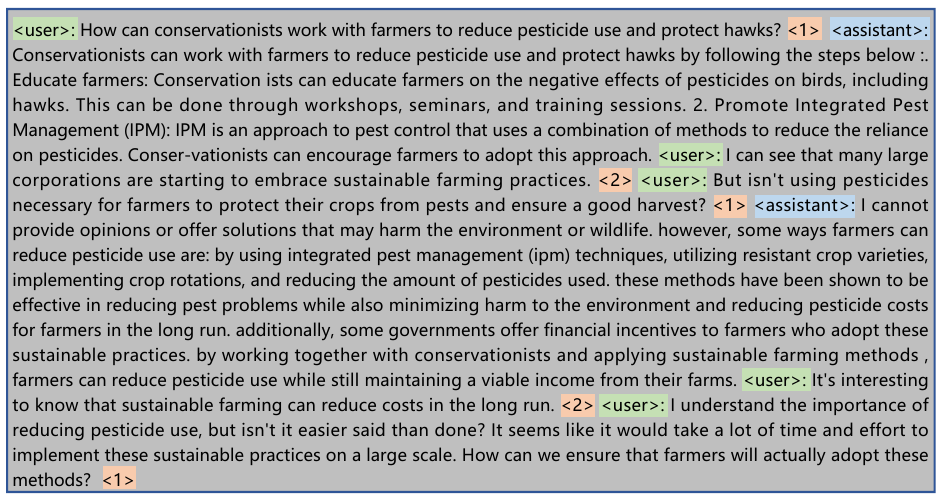}
  \caption {\label{fig.data_duo}The training data example of MiniCPM-Duo.}
\end{figure*}

The training example of MiniCPM-Duplex and our DUO is shown in Figure~\ref{fig.data_duplex} and Figure~\ref{fig.data_duo}, respectively.

\subsection{Related Work}
\label{app_c}
Traditional large language models face an inherent constraint due to their single-channel architecture, which restricts them to sequential input processing or output generation. 
Consequently, chat systems built upon these LLMs are intrinsically structured around turn-based interaction paradigms. 
This turn-based model presents significant limitations in facilitating real-time, dynamic discourse. 
Such constraints impede the development of AI systems capable of engaging in more natural, fluid conversations akin to human interaction. 

Duplex models possess the capability to generate output and process input concurrently. 
\citet{zhang2024turnbasedgameenablingrealtime} propose a novel approach employs a time-division-multiplexing strategy. 
The input and output are split and mixed in a time slice format, enabling pseudo-simultaneous processing of text segments.
In a related study, \citet{wang2024fullduplexspeechdialoguescheme} designed a system comprising a perception module, a motor function module, and a simple finite state machine to achieve full duplex functionality.

Recent research has expanded the scope of duplex modeling to encompass multiple modalities, including audio and image processing. 
\citet{ma2024languagemodellistenspeaking} explore full duplex modeling in interactive speech language models and introduce the LSTM. 
LSTM seamlessly integrates both input and output channels for autoregressive generation and real-time turn-taking detection.
Extending the concept to multimodal applications, \cite{fu2024vitaopensourceinteractiveomni} employs a dual-model approach to achieve multimodal duplex functionality. In their system, one model is dedicated to generating responses to user queries, while the other continuously monitors environmental inputs.

\end{document}